\documentclass{article}

\PassOptionsToPackage{numbers, compress}{natbib}
 \usepackage[dblblindworkshop, final]{neurips_2025}

\usepackage[utf8]{inputenc} % allow utf-8 input
\usepackage[T1]{fontenc}    % use 8-bit T1 fonts
\usepackage{hyperref}       % hyperlinks
\usepackage{url}            % simple URL typesetting
\usepackage{booktabs}       % professional-quality tables
\usepackage{amsfonts}       % blackboard math symbols
\usepackage{nicefrac}       % compact symbols for 1/2, etc.
\usepackage{microtype}      % microtypography
\usepackage{xcolor}         % colors
\usepackage{todonotes}      % todos
\usepackage{wrapfig}        % wrap text around figures
\usepackage{lipsum}         % for filler text
\usepackage{float}          % for figure placement

\title{Multi-Modal AI for Remote Patient Monitoring in Cancer Care}

\workshoptitle{Multimodal Representation Learning for Healthcare}

\makeatletter
\newlength{\halotitlelift}
\renewcommand{\@toptitlebar}{%
    \hrule height 4\p@\relax
    \vskip 0.12in
    \vskip -\parskip
}
\renewcommand{\@bottomtitlebar}{%
    \vskip 0.18in
    \vskip -\parskip
    \hrule height 1\p@\relax
    \vskip -0.1in
}
\renewcommand{\@maketitle}{%
    \vbox{%
        \hsize\textwidth
        \linewidth\hsize
        \vskip -\halotitlelift
        \@toptitlebar
        \centering
        {\LARGE\bf \@title\par}
        \@bottomtitlebar
        \if@anonymous
            \begin{tabular}[t]{c}\bf\rule{\z@}{24\p@}
                Anonymous Author(s) \\
                Affiliation \\
                Address \\
                	exttt{email} \\
            \end{tabular}%
        \else
            \def\And{%
                \end{tabular}\hfil\linebreak[0]\hfil%
                \begin{tabular}[t]{c}\bf\rule{\z@}{24\p@}\ignorespaces%
            }
            \def\AND{%
                \end{tabular}\hfil\linebreak[4]\hfil%
                \begin{tabular}[t]{c}\bf\rule{\z@}{24\p@}\ignorespaces%
            }
            \begin{tabular}[t]{c}\bf\rule{\z@}{24\p@}\@author\end{tabular}%
        \fi
        \vskip 0.1in \@minus 0.08in
    }
}
\makeatother

\author{
  Yansong Liu\textsuperscript{1}\quad
  Ronnie Stafford\textsuperscript{1,2}\quad
  Pramit Khetrapal\textsuperscript{1,2}\quad
  Huriye Kocadag\textsuperscript{1}\\
  \textbf{Gra\c{c}a Carvalho\textsuperscript{1,2,3}}\quad
  \textbf{Patricia de Winter\textsuperscript{1}}\quad
  \textbf{Maryam Imran\textsuperscript{1}}\quad
  \textbf{Amelia Snook\textsuperscript{1}}\\
  \textbf{Adamos Hadjivasiliou\textsuperscript{1}}\quad
  \textbf{D. Vijay Anand.\textsuperscript{1}}\quad
  \textbf{Weining Lin\textsuperscript{1}}\quad\\
  \textbf{John Kelly\textsuperscript{1,2}}\quad
  \textbf{Yukun Zhou\textsuperscript{1}}\quad
  \vspace{3pt}
  \textbf{Ivana Drobnjak\textsuperscript{1}}\\
  \texttt{\{yansong.liu.18, r.stafford.12, p.khetrapal, h.kocadag,}\\
  \texttt{graca.carvalho, p.winter, maryam.imran.20, amelia.snook.20,}\\
  \texttt{adamos.hadjivasiliou.17, v.dharmalingam, w.lin.16}\\
  \vspace{3pt}
  \texttt{\j.d.kelly, yukun.zhou.19, i.drobnjak\} @ucl.ac.uk}\\
  \textsuperscript{1}University College London, UK, WC1E 6BT\\
  \textsuperscript{2}Ethera Health LTD, 125 Wood Street, London, UK, EC2V 7AW\\
  \textsuperscript{3}Centro Algoritmi, Universidade do Minho, Braga, Portugal\\
}

\begin{document}

\maketitle

% \begingroup
% \renewcommand\thefootnote{}
% \footnotetext{\raggedright%
%   Correspondence: University College London: \texttt{\{yansong.liu.18, r.stafford.12, p.khetrapal, h.kocadag, graca.carvalho, p.winter, maryam.imran.20, amelia.snook.20, adamos.hadjivasiliou.17, v.dharmalingam, w.lin.16, j.d.kelly, yukun.zhou.19, i.drobnjak\}@ucl.ac.uk}. Ethera Health: \texttt{\{ronnie.stafford, huriye.kocadag\}@ethera.health}.%
% }
% \addtocounter{footnote}{-1}
% \endgroup

\begin{abstract}
  For patients undergoing systemic cancer therapy, the time between clinic visits is full of uncertainties and risks of unmonitored side effects. To bridge this gap in care, we developed and prospectively trialed a multi-modal AI framework for remote patient monitoring (RPM). This system integrates multi-modal data from the HALO-X platform, such as demographics, wearable sensors, daily surveys, and clinical events. Our observational trial is one of the largest of its kind and has collected over 2.1 million data points (6,080 patient-days) of monitoring from 84 patients. We developed and adapted a multi-modal AI model to handle the asynchronous and incomplete nature of real-world RPM data, forecasting a continuous risk of future adverse events. The model achieved an accuracy of 83.9\% (AUROC=0.70). Notably, the model identified previous treatments, wellness check-ins, and daily maximum heart rate as key predictive features. A case study demonstrated the model's ability to provide early warnings by outputting escalating risk profiles prior to the event. This work establishes the feasibility of multi-modal AI RPM for cancer care and offers a path toward more proactive patient support. \href{https://github.com/LiuYYSS/EurIPS2025}{https://github.com/LiuYYSS/EurIPS2025}
\end{abstract}
\vspace{-10pt}

\section{Introduction}
For individuals undergoing cancer systemic therapy (such as chemotherapy), the period between clinical appointments unfolds at home and is often fraught with uncertainty and the risk of unmonitored side effects or rapid health deterioration~\cite{oakley2017neutropenic}. In the meantime, due to physical distance, clinicians have limited visibility into their patients' at-home daily well-being and the subtle onset of treatment-related side effects, creating a critical gap in cancer care~\cite{chen2023symptom}.

RPM has the potential to address this problem by using multiple sources (such as wearable sensors, mobile apps, and other digital tools) to gather at-home health data~\cite{rockey2025effect}. To unlock the true potential of RPM, multi-modal AI was involved to fuse sources together to generate actionable decisions~\cite{soenksen2022integrated}. This approach is highly likely to fill the gap because it mirrors the multi-aspect reasoning process of clinicians at home environments. 

Previous research in multi-modal AI RPM in cancer care handled the multi-modalities poorly. Some recent research was based on single modality (wearable only~\cite{cay_harnessing_2024, low2025consumer}, survey only~\cite{kairis2025daily, rockey2025effect}). Some research had to reduce sampling frequency because certain modality was sampled slower than the others~\cite{jacobsen2023wearable, liu2023evaluating}. There is a lack of standardised framework to address this.

This study presents our observational prospective trial using a framework that collects and uses multi-modal RPM data to forecast adverse clinical events. Specifically, we explored the asynchronous and non-random missingness problems through a token-based transformer, validated on the trial data. Ultimately, this work establishes the feasibility of multi-modal AI RPM for cancer care and offers a path toward more proactive patient support.

%\newpage
\section{Methods}
\begin{figure}[t]
    \centering
    % \vspace{-10pt}
    \includegraphics[width=0.85\textwidth]{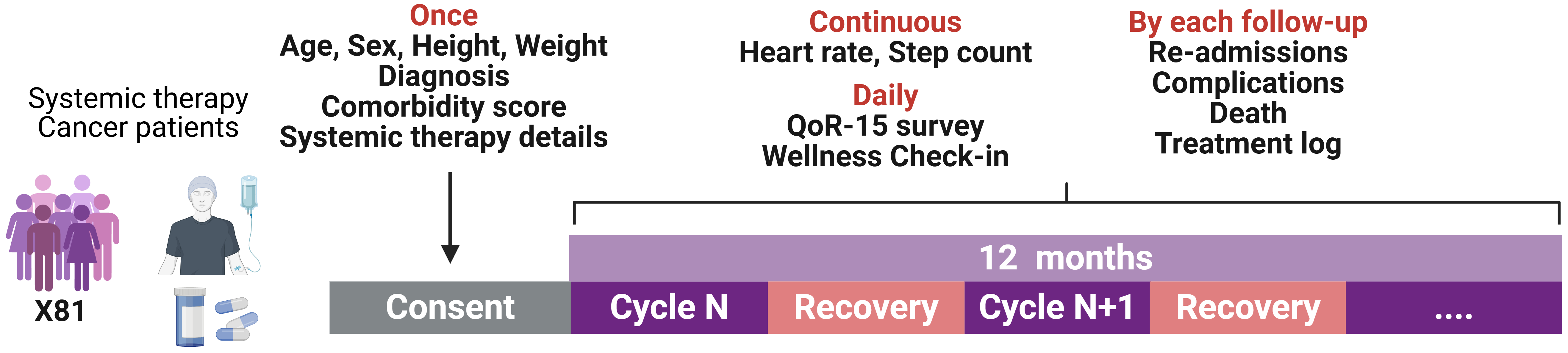}
    \caption{Overview of trial design. Multimodal RPM data collection up to 12 months}
    \vspace{-10pt}
    \label{fig:data_collection}
\end{figure}

\begin{wrapfigure}{r}{0.42\textwidth}
  \vspace{-40pt}
  \centering
  \includegraphics[width=0.40\textwidth]{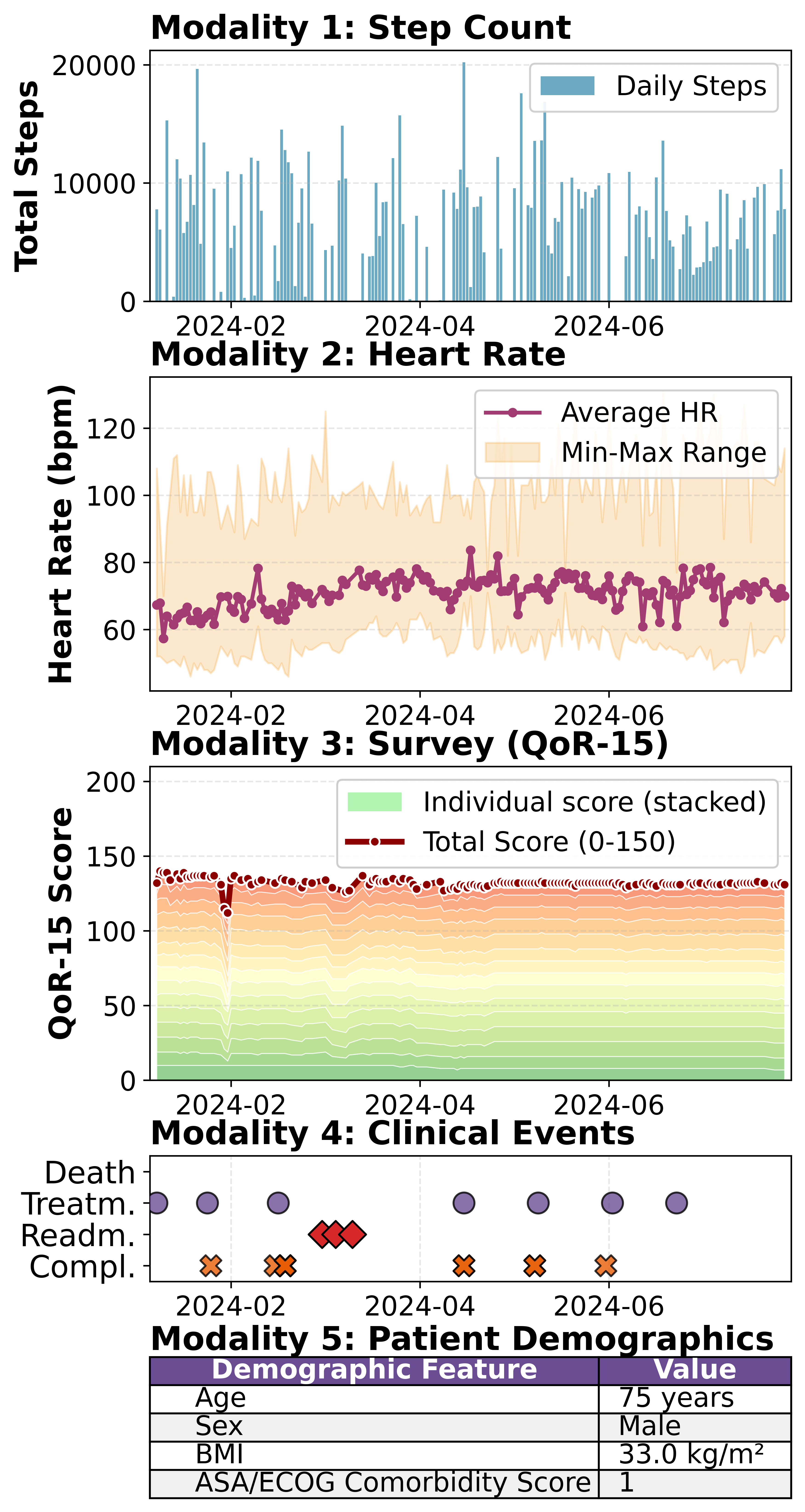}
  \caption{Representative patient timeline illustrating longitudinal asynchronous multi-modalities data with missingness. Raw sensor data were aggregated to daily values to avoid clustering in this visualisation. Treatnm.=Treatment; Readm.=Re-admission; Compl.=Complications.}
  \label{fig:multimodal_viz}
  \vspace{-10pt}
\end{wrapfigure}

\paragraph{Multi-modal RPM data Collection:}  A prospective observational trial (IRAS 312296) was designed based on the HALO-X multi-modal data collection platform \cite{halox2025infra}. The trial aimed at testing the feasibility of the HALO-X platform in patients with advanced cancer undergoing systemic therapy. Patient recruitment started in March 2023 (inclusion \& exclusion criteria see appendix \ref{appendix:inclusion-exclusion}). Detailed collection process is illustrated in Figure \ref{fig:data_collection}. Multi-modal data was collected and combined via the platform: baseline characteristics, wearable data in 5-minute epochs, surveys, and clinical events. Surveys used were the 15-item Quality of Recovery questionnaire (QoR-15, 0-10 scale per item, 10=best)~\cite{stark2013development} and wellness check-in (an optional yes/no question to indicate overall health today). Treatment log contains information from treatment events such as treatment type (chemotherapy, hormone therapy, immunotherapy etc.), health-caused dose reduction or treatment delay (e.g. low blood platelets).

\begin{figure}[t]
    \centering
    % \vspace{-10pt}
    \includegraphics[width=0.9\textwidth]{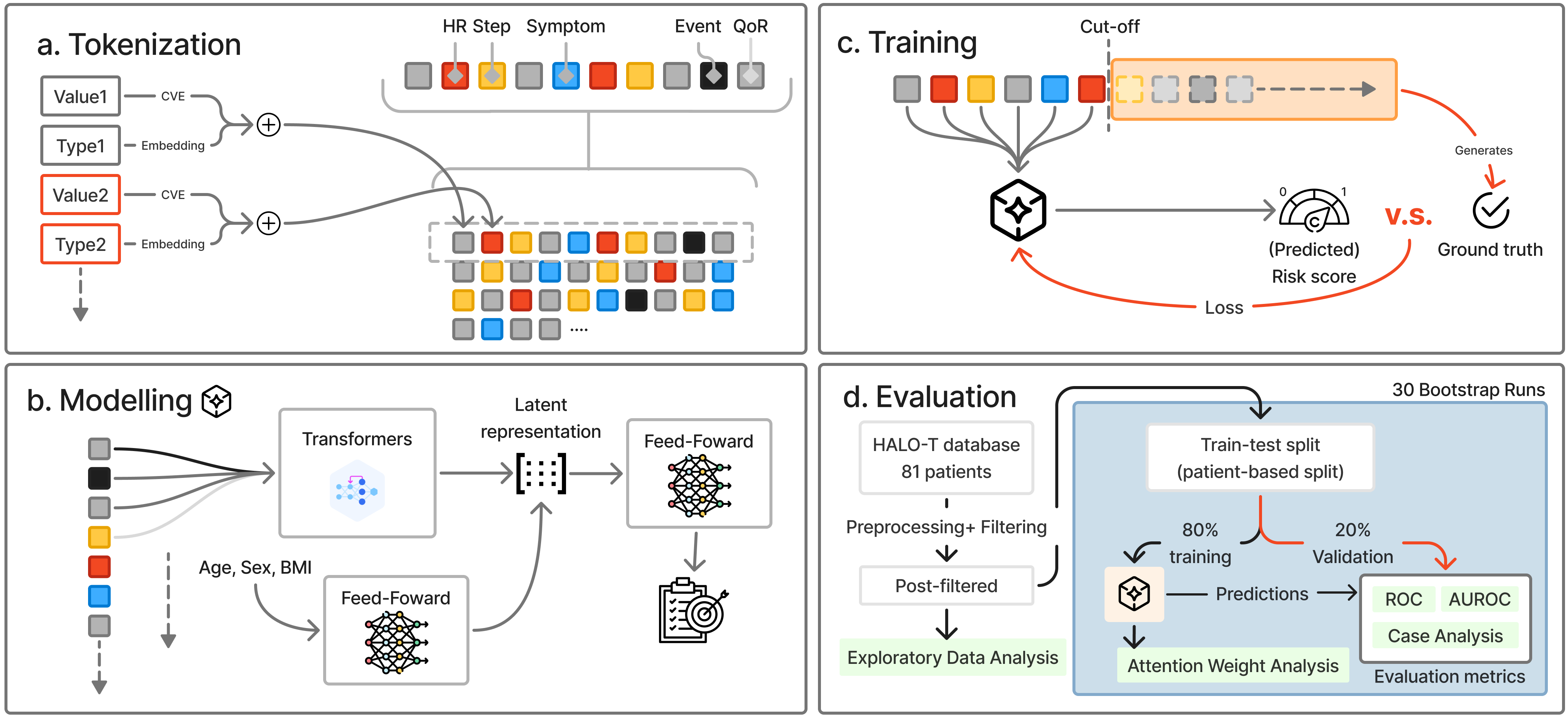}
    \caption{Model architecture and training pipeline. (a) Value and type of each observation were extracted and tokenized, CVE=Continuous Variable Encoder; (b) Transformer layers for temporal modelling. Static features were encoded via a separate feed-forward neural network; (c) Model uses data before cut-off to generate a binary classification forecast; (d) Evaluation paradigm.}
    \label{fig:model_development}
    \vspace{-10pt}
\end{figure}

\paragraph{Multi-modal RPM Model Development:} Data stream asynchronicity is a key consideration when building multi-modal AI for RPM data. The problem appears as: wearable sensors sampled every 5 minutes, survey responses arrived daily, and clinical events occurred irregularly across months of treatments. The way that traditional supervised learning model works either causes the least frequent feature to be re-used numerous times or binning the most frequent feature to a lower frequency (which significantly reduces temporal resolution). Additionally, patients with poor health are known to use monitoring services less \cite{McClaine2024, goldberg2021data, Ortiz2024}. This introduces missingness not at random (MNAR). MNAR in healthcare is notorious for prevalence and tricky to solve~\cite{little2002statistical,sterne2009multiple}. To develop an AI model tailored to these data characteristics, we took the concept and adapted a transformer-architecture from the STraTS~\cite{tipirneni2022self} with some key adjustments. First, tokenizer encodes each observation as an independent token, creating token sequences where each modality contributes tokens at its native sampling rate, avoiding the first problem mentioned. Second, because each observation is tokenized independently, missing observations can be simply skipped rather than forced to be imputed, avoiding the second problem mentioned. Third, we explicitly engineered missingness features (daily device wearing percentage and continuous absence duration) as additional tokens to enhance missingness learning. The model architecture and training/evaluation pipeline is illustrated in Figure \ref{fig:model_development}.

The model was trained to forecast future adverse clinical events (general practitioner visit due to treatment-related reasons, accident\&emergency department visit, re-admission, treatment delay/dose reduction, or death) within rolling 4-week windows. The output is a float value between 0 and 1, where 1 represents extremely high risk. The model takes all available historical monitoring data (minimum 2 weeks) as input. Prior to modelling, filtering rules (Appendix table \ref{tab:patient-filtering}) were applied to remove patients with artifacts. Heart rate was clipped to 40-200 bpm, steps clipped to 0-600 per 5-minute epoch. Training employed sliding window re-sampling to generate multiple samples from each patient's longitudinal trajectory, with patient-level stratified splitting (80-20) to prevent data leakage. To reduce the stress on GPU memory and improve learning efficiency, raw sensor data were aggregated into representative daily values and max sequence length was kept at 1000 tokens. The hyperparameters were selected manually based on the original hyperparameters from STraTS with batch size set to 128, 80 epochs, and 5e-4 learning rate. During evaluation, bootstrap aggregation with 30 iterations was used to provide performance estimates with confidence intervals. Attention weight was extracted from the attention layer for feature importance analysis. Evaluation results from each sliding window over time were put together and connected to generate risk trajectory over time for case-analysis. A single NVIDIA RTX GPU was used, and took 2 hours to complete all bootstraps.

\section{Results}

\begin{wrapfigure}{r}{0.37\textwidth}
  \vspace{-10pt}
  \centering
  \includegraphics[width=0.35\textwidth]{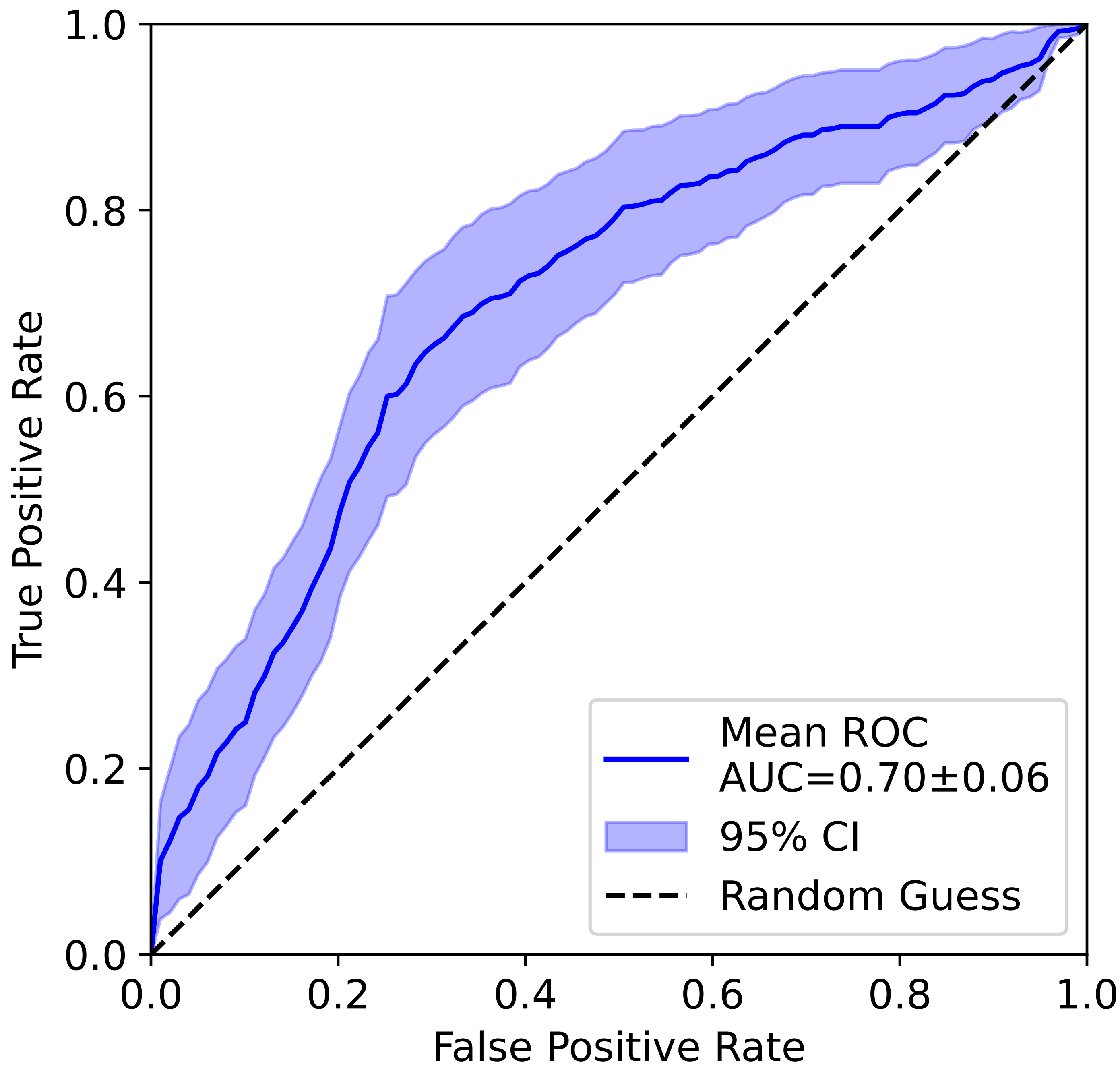}
  \caption{ROC curve of the model with 95\% CI.}
  \label{fig:roc}
  \vspace{-10pt}
\end{wrapfigure}

Up to December 2024 (the date the dataset was extracted), the HALO-X platform collected over 2.1 million data points (top three sources: 1.2M in heart rate, 0.5M in step count, 0.07M in QoR-15) from 6,080 patient-days of monitoring data, marking one of the largest prospective trial datasets in multi-modal RPM in cancer systemic therapy to date. 50 patients (5,296 patient-days, 87\%) were included in the final analysis after filtering (Appendix table \ref{tab:patient-filtering}). On average, each patient was monitored for 98 days (median 76 days, max 298 days, min 42 days). 30 features covering multiple modalities were extracted for training and prediction (Appendix table \ref{tab:list-of-features}). Among 50 patients, 17 patients produced 66 adverse events during the monitoring period (treatment delay/dose reduction was the most frequent adverse event, accounting for 32 occurrences, 48\%). Because only events within 4-week windows were considered as positive labels, among 9,313 sliding windows generated, only 1,265 (13.6\%) were labelled positive. The dataset is moderately unbalanced with a 6.4:1 negative-to-positive ratio.

\begin{figure}[htbp]
    % \vspace{-10pt}
    \centering
    \includegraphics[width=0.98\textwidth]{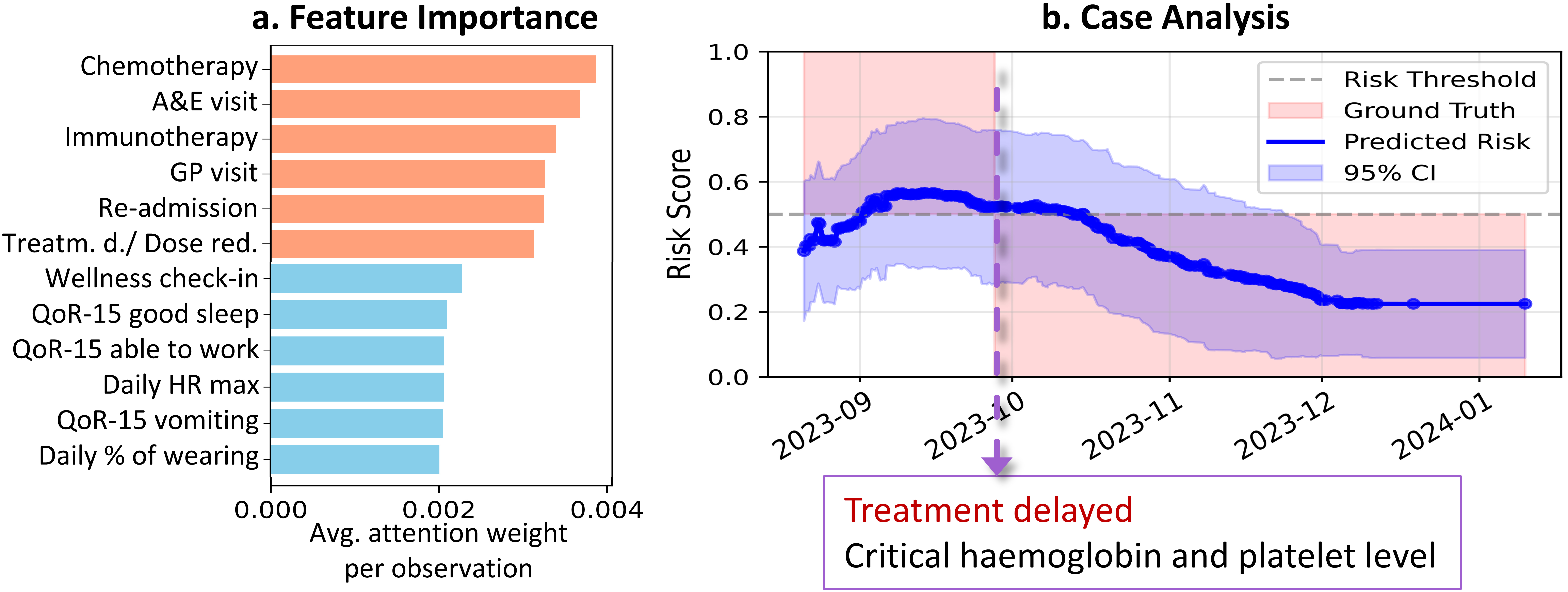}
    \caption{(a) Overview of the most important features, orange bars indicate event features which are less frequent but impactful. Blue bars are remote monitoring feature which happens more regularly and indicates recovery status; (b) Example patient trajectory over time with clinical events annotated. The risk score (y-axis) ranges from 0 to 1, where 1 indicates extremely high risk. Treatm. d./Dose Red.=Treatment Delay/Dose Reduction; A\&E=Accident \& Emergency department; GP=General Practitioner; QoR-15=Quality of Recovery 15 survey.}
    \label{fig:combined_feature_importance_and_risk_trajectory}
    \vspace{-10pt}
\end{figure}

The model achieved 0.70 (95\% CI: 0.64-0.76) area under ROC and 83.9\% (95\% CI: 81.5-86.3\%) accuracy (figure \ref{fig:roc}). Ranking the extracted attention weights by importance (figure \ref{fig:combined_feature_importance_and_risk_trajectory} a and b), features in the clinical event type are generally stronger than remote monitoring features, likely due to lower frequency, thus higher information density. The strongest feature in event type is previous chemotherapy treatment, followed by A\&E department visit and immunotherapy treatment. The top remote monitoring feature is wellness check-in, followed by good sleep and able to work (two items in QoR-15). The top wearable sensor feature is daily maximum heart rate. Percentage of daily wearing is also among the top features, indicating the model learned missingness information. In figure \ref{fig:combined_feature_importance_and_risk_trajectory} c, we showcase a representative patient risk trajectory over time with clinical events annotated. The patient was selected from the testing set on which the model had never been trained. The predicted risk elevated ahead of the treatment delay event. Even though haemoglobin and platelet level were not among the features, the model was able to capture the subtle health deterioration via other data. After about a month of recovery, the patient received the next treatment with intended dosage, and the risk score dropped back to a low level. This case demonstrates the potential of the model in identifying high-risk periods and comparing recovery trajectories.

\section{Discussion}
This study shows that a multi-modal AI remote patient monitoring (RPM) in cancer therapy is feasible. We ran a prospective trial that collected >2.1 million data points across 6,000 patient-days, and developed and adapted an algorithm that addresses asynchronous data streams and non-random missingness (MNAR) common in real-world monitoring. The model's initial performance (AUC = 0.70, accuracy = 84\%) is solid, but the key contribution is the practical and scientific insight: combining continuous wearable signals, daily surveys, and clinical event history produces a dynamic risk profile that better captures patient state between clinic visits.

Our exploratory analyses revealed which inputs drive prediction. Clinical event history holds strong predictive value. Among remote monitoring signals, our optional wellness check-ins, sleep quality, able to work, and daily maximum heart rate emerged as important indicators, showing the value of collecting daily well-being measurements. A case study shows a rising risk score before a treatment delay, illustrating how this multi-modal AI RPM can be used for early warning signs and could enable proactive clinical intervention. Overall, this work demonstrates feasibility, identifies high-value features for RPM in cancer care, and provides a roadmap for larger validation and clinical integration.

As an analysis of an ongoing study, this work has several limitations. The biggest being data size, and a series of other limitations caused by this. The analysis was based on longitudinal data from just 50 patients. With that said, given how new this field is and the difficulty of collecting prospective data in healthcare, the dataset is already one of the largest prospective trials in multi-modal RPM in cancer systemic therapy to date. Future work should focus on expanding the cohort, validating across multiple centres, and improving accuracy. This study serves as a crucial first step, providing the blueprint and initial evidence needed to build toward a future where multi-modal AI and remote monitoring become integral tools for delivering more personalized and proactive cancer care.

\newpage
\begin{ack}
We would like to thank Ethera Health LTD for the support given to this work through the research collaboration, \textit{Proof of Concept Study to Test a Framework for Integration of Data Streams to Explore the Relation between Measures of Molecular Genomic Markers and Performance in Patients Receiving Chemotherapy for Metastatic Cancer} and the use of the HALO-X RPM data collection platform.

We would also like to acknowledge the support of the Urology Foundation, the Champniss Foundation and the Champniss Charitable Trust through the grant, \textit{Testing The Relation Between The Host Immune Response And Activity In Patients Receiving Chemotherapy For Solid Cancers}.
\end{ack}

\bibliographystyle{unsrtnat}
\bibliography{references}

%%%%%%%%%%%%%%%%%%%%%%%%%%%%%%%%%%%%%%%%%%%%%%%%%%%%%%%%%%%%

\newpage
\appendix

\section{Technical Appendices and Supplementary Material}

\subsection{Inclusion and Exclusion Criteria}\label{appendix:inclusion-exclusion}
\begin{itemize}
    \item Inclusion criteria:
    \begin{itemize}
        \item Diagnosis of advanced or haematological cancer undergoing systemic therapy;
        \item Planned for at least one cycle of standard-of-care systemic therapy;
        \item Ability to provide informed consent;
        \item Ambulatory without assistance or walking aids;
        \item Possession of an Android or iOS smartphone with willingness to use the HALO-X monitoring smartphone application.
    \end{itemize}
    \item Exclusion criteria:
    \begin{itemize}
        \item Physical disabilities precluding daily walking;
        \item Inability to provide informed consent;
        \item Inability to operate the HALO-X monitoring smartphone application;
        \item Medical or psychiatric conditions that would affect study completion, as determined by the investigator.
    \end{itemize}
\end{itemize}

\subsection{Data Filtering and Feature Overview}
\begin{table}[H]
\centering
\resizebox{\columnwidth}{!}{%
\begin{tabular}{lll}
    & Filtering reason                                           & Num. Patient Removed \\
1 & Patients that were not followed up for clinical events     & 4                    \\
2 & Patients with empty record                                 & 3                    \\
3 & Patients that were not allocated with a device             & 3                    \\
4 & Patients with no treatment received in monitoring period   & 5                    \\
5 & Patients providing any RPM data for \textless 3 days         & 11                   \\
6 & Patient involved in a car accident                         & 1                    \\
7 & Patients received stem cell therapy (distinctive pattern and limited population) & 5 \\
8 & Patients who joined the trial too shortly before data extraction & 3                    \\ \hline
    & \multicolumn{1}{r}{Population extracted}                   & 85                   \\
    & \multicolumn{1}{r}{Patients Affected}                      & 35                   \\
    & \multicolumn{1}{r}{Final population}                       & 50                  
\end{tabular}%
}
\caption{Patient filtering rules, number of patients affected by each rule, and total number of patients remaining after filtering.}
\label{tab:patient-filtering}
\end{table}

\begin{table}[H]
\resizebox{\columnwidth}{!}{%
\begin{tabular}{l|llll}
\hline
Categories & Wearable                       & Survey                  & Previous Clinical Events                       & Demographics \\ \hline
           & Daily max HR                   & QoR-15 total score      & Chemotherapy                                   & Age          \\
           & Daily total steps              & QoR-15 individual score x15 & Hormone therapy                                & Gender       \\
           & Daily \% of wearing            & Complication seriousness     & Immunotherapy                                  & BMI          \\
           & Duration of continuous absence & Wellness check-in       & Mixed therapy                                  &              \\
           &                                &                         & Re-admission                                   &              \\
           &                                &                         & General practitioner visit                     &              \\
           &                                &                         & Accident \& Emergency department visit         &              \\
           &                                &                         & Dose reduction / Treatment delay due to health &              \\ \hline
\end{tabular}%
}
\caption{Overview of features used in the model. BMI=Body Mass Index; HR=Heart Rate; QoR=Quality of Recovery.}
\label{tab:list-of-features}
\end{table}

\end{document}